\documentclass[10pt,twocolumn]{article}

\usepackage{amsmath}
\usepackage{amssymb}
\usepackage{graphicx}
\usepackage{cite}
\usepackage{xcolor}
\usepackage[colorlinks=true, linkcolor=blue, citecolor=blue, urlcolor=blue]{hyperref}
\usepackage{subcaption}
\usepackage{algorithm}
\usepackage{algpseudocode}
\usepackage{booktabs}
\usepackage{array}
\usepackage{float}
\usepackage{minted}
\usepackage{tikz}
\usepackage{listings}
\usepackage{xspace}
\usepackage{cleveref}

\newcommand{\EA}{\textit{EA}\xspace}
\newcommand{\WFC}{\textit{WFC}\xspace}
\newcommand{\PSO}{\textit{PSO}\xspace}
\newcommand{\GWO}{\textit{GWO}\xspace}
\newcommand{\MapElites}{\textit{MAP-Elites}\xspace}
\newcommand{\Duckietown}{\textit{Duckietown}\xspace}

\title{Elite Lanes: Evolutionary Generation of Realistic Small-Scale Road Networks\\\large{Preprint. Work has been accepted for GECCO 2026 as poster.}}

\author{
Artur Morys-Magiera\\
AGH University of Krakow, Poland\\
\texttt{amorys@agh.edu.pl}
\and
Marek Długosz\\
AGH University of Krakow, Poland\\
\texttt{mdlugosz@agh.edu.pl}
\and
Paweł Skruch\\
AGH University of Krakow, Poland\\
\texttt{pawel.skruch@agh.edu.pl}
}

\date{} 

\begin{document}

\maketitle

\begin{abstract}
  We present a comparative study of methods for generating realistic, constrained small- to medium-scale road networks with built-in redundancy. In this research, we evaluate the proposed Evolutionary Algorithm (EA) with connectivity and redundancy constraints against the Wave Function Collapse (WFC) method - commonly used in procedural terrain generation for games - and swarm algorithms: Particle Swarm (PSO) and Gray Wolf (GWO). Our focus is on producing realistic, redundant road networks suitable for vision, localization and navigation problems. We evaluate metrics: connectivity, cycles, intersections, dead ends, graph cut-edges while enforcing physical plausibility. We propose an EA and its extended version with elitism via MAP-Elites method. We detail the implementation, constraints, metrics and provide both visual and quantitative comparisons with baselines. Results highlight how fitness function design choices affect the structural characteristics of generated networks and highlight the impact of specific constraints in practical applications. Our contribution is a method for creating realistic synthetic datasets from sparse tile definitions derived from real-world data. We demonstrate a practical application by generating realistic maps using a laboratory-collected tileset from a Duckietown city model. Our approach performs coherent geometric transformations on metadata, in this work exemplified by semantic segmentation masks of the generated road networks.
\end{abstract}

\maketitle


\section{Introduction}
\label{sec:introduction}
The generation of realistic synthetic datasets is a critical challenge across multiple domains, including robotics, autonomous vehicle simulation, video game design, and urban planning. While it is often possible to generate synthetic datasets, numerous issues persist. A key challenge addressed in this research is the \textit{reality gap}, which refers to the phenomenon where models trained on synthetic data frequently fail to transfer learned knowledge effectively to real-world data. The phenomenon has been described by Steinhoff et al. \cite{steinhoff2025simulation} for synthetic data in general, and by Duc et al. \cite{duc2025mind} specifically in the context of autonomous robotics.

Researchers have proposed several approaches to mitigate this issue. The most straightforward strategy - combining synthetic and real-world data to bridge the gap - has been widely employed, for example by Khose et al. \cite{khose2024skyscenes}. Zhao et al. \cite{zhao2025sim} suggested feature-level adaptation, aligning the feature distributions between synthetic and real domains to improve transferability. Liao et al. \cite{liao2026bridging} focused on enhancing the visual fidelity of generated datasets by improving rendering quality to produce more realistic data.

In this study, the authors focus on the aspect of generating realistic, constrained complete road networks with stop lines and dashed lane markings to train computer vision models for semantic segmentation across four classes: background, roads, stop lines, and lane separator lines. A major constraint in this work is the limited availability of real-world data, leading to a low-data scenario. Specifically, the authors assume that only a small number of samples of complete road networks can be obtained, reflecting practical limitations such as scarce resources for mapping real environments or the labor-intensive process of manually creating, photographing, and labeling large numbers of networks. Generated datasets of this type have applications in:
\begin{enumerate}
    \item \textbf{Training datasets for perception:} Vision-based localization and navigation models require diverse road network configurations to generalize well.
    \item \textbf{Simulation environments:} Autonomous vehicle and robot navigation systems need varied, realistic environments for training and testing.
    \item \textbf{Navigation benchmark diversity:} Systematic variation of network topology enables controlled evaluation of navigation algorithms.
\end{enumerate}

Therefore, this work focuses on generating synthetic datasets of realistic road networks for small-scale environments. The research was experimentally tested on the laboratory-grade robotics platform \Duckietown \cite{paull2017duckietown}.

Approaches to road network or general environment models generation in literature fall into the following categories:
\begin{itemize}
    \item \textbf{Constraint satisfaction:} Methods like Wave Function Collapse (WFC) that enforce local constraints. WFC has been widely applied for game content generation, as described by Kim et al. \cite{kim2019automatic}, as well as for a domain much closer to the one described in this article: parking layouts, as described by Lan et al. \cite{lan2023underground}.
    \item \textbf{Swarm intelligence:} Population-based optimization, such as the Particle Swarm Optimization algorithm successfully applied by Cipriani et al. \cite{cipriani2020particle} for transit network design.
    \item \textbf{Evolutionary algorithms:} Direct search in the space of road network topologies.
\end{itemize}

This paper contributes a systematic comparison of these approaches with a proposed solution for training an \EA with \MapElites \cite{mouret2015illuminating} for quality-diversity optimization. The \MapElites method, described by Mouret et al., maintains an archive of non-dominated solutions distributed across the behavior space, enabling discovery of networks with diverse structural characteristics while maintaining solution quality. As no implementations for the WFC algorithm or the PSO algorithm have been found associated with existing research, baseline implementations were created by the authors.

The primary contributions of this work are:
\begin{enumerate}
    \item \textbf{Systematic comparison}: Evaluation of \WFC, \PSO, \GWO, and proposed configurations of \EA and \MapElites against quantitative metrics.
    
    \item \textbf{MAP-Elites application}: Demonstration that quality-diversity optimization using \MapElites outperforms all other approaches and is followed by \EA which is superior to \WFC, \PSO and \GWO, especially in terms of diversity of solutions. An implied contribution is presentation of the fitness function composition.
    
    \item \textbf{Real-world application}: Practical generation of synthetic \Duckietown maps with coherently transformed binary masks, constituting a synthetic road network semantic segmentation dataset. The approach can be easily adjusted to match the specific amount and characteristics of markings / entities in other cases.
\end{enumerate}

\section{Problem Formulation}
\label{sec:problem}

\subsection{Tile-Based Road Network Representation}
\label{subsec:tile_representation}

We represent road networks as a grid of square tiles, each encoding directional connectivity information. Each tile is encoded as a 4-bit integer representing connections in four cardinal directions (in order): North, East, South, and West (NESW).

\begin{equation}
\text{tile} = 2^3 \cdot b_N + 2^2 \cdot b_E + 2^1 \cdot b_S + 2^0 \cdot b_W
\end{equation}

where $b_N, b_E, b_S, b_W \in \{0, 1\}$ indicate whether the tile has a road connection in that direction.

\subsection{Constraints}
\label{subsec:constraints}

The generation process must satisfy multiple hard and soft constraints:

\begin{enumerate}
    \item \textbf{Connectivity matching:} If tile $t_1$ connects in direction $d$ to tile $t_2$, then $t_2$ must connect back in the opposite direction $\bar{d}$.
    
    \item \textbf{Boundary constraints and dangling ends:} Tiles on the grid boundary cannot have outward connections. Moreover, in a perfect case, the graph shall have no leaves, in which case there are no "dead ends" in the roads. This means penalization of the number of leaves in the graph.
    
    \item \textbf{Crossing adjacency constraint:} In a real-world scenario, although it is not impossible to have two adjacent crossings (defined as tiles with 3 or more connections) interconnected directly, it is undesirable. Therefore, adjacent crossings shall be penalized.
    
    \item \textbf{Single graph:} To generate a network of roads and not a set of different networks, the result shall include a single, connected graph.

    \item \textbf{Traffic balance}: Real road networks are designed to include redundancy such that critical paths (unique edges of the graph that connect two arbitrary nodes) are rare, to balance the traffic. Therefore, the existence of graph cut-edges (bridges) shall be minimized and at the same time, the count of cycles shall positively affect the quality indicator of a network.
\end{enumerate}

\subsection{Evaluation Metrics}
\label{subsec:metrics}

We employ the following metrics to capture different aspects of network realism:

\begin{enumerate}
    \item \textbf{Connected components}: the amount of graphs in the result. This metric is calculated using a Depth-First Search (DFS) algorithm counting connected components.
    
    \item \textbf{Cyclomatic complexity}: The cyclomatic complexity \cite{mccabe1976complexity} of the network graph, computed as $M = E - N + P$, where $E$ is edges, $N$ is nodes, and $P$ is connected components. This metric has been chosen instead of just the cycle count to account for the branching structure of the network.

    \item \textbf{Straight roads}: Since straight road tiles have a $2$-way connectivity, statistically it is easier to place other types of tiles (such as crossings) that have a higher connectivity and therefore fit more flexibly. Therefore, consecutive sequences of directional tiles (horizontal or vertical) are scored quadratically to reward longer runs.
    
    \item \textbf{Adjacent crossing violations}: Count of tiles with 3 or more connections that are placed next to each other, which constitutes for an improbable scenario. The count is scaled by the amount of edges.
    
    \item \textbf{Dangling ends (dead ends)}: Tiles with exactly one connection (unfavorable).
    
    \item \textbf{Graph cut-edges}: Edges required to disconnect the graph into separate components (graph connectivity measure). This entity is unfavorable.
    
    \item \textbf{Chained turns}: To prevent chaining turns ("zig-zag" turns), the count of the turns is used to penalize a solution.
    
    \item \textbf{Coverage}: The normalized ratio of area with placed tiles to covered tiles.
    
    \item \textbf{Straight roads}: to promote roads, the length of consecutive straight roads is squared and added as a bonus.
\end{enumerate}

\section{Related Work}
\label{sec:related}

\subsection{Procedural Content Generation}
\label{subsec:pcg}

Procedural content generation (PCG) in games and simulations has a long history. Early approaches used noise-based methods (Perlin noise, Simplex noise) \cite{lagae2010survey} for generation of constructs. More recent approaches employ constraint satisfaction and quality-diversity optimization.

\subsection{Wave Function Collapse}
\label{subsec:wfc_related}

Wave Function Collapse is a constraint satisfaction algorithm inspired by quantum mechanics \cite{kim2019automatic}. It has been used in procedural game design for texture synthesis and map generation. \WFC works through:
\begin{enumerate}
    \item Superposition initialization: all cells can be any tile type.
    \item Entropy-based collapse: collapse the lowest-entropy cell.
    \item Constraint propagation: update neighbor possibilities.
    \item Iteration: repeat until fully collapsed or contradiction.
\end{enumerate}

\subsection{Swarm Intelligence}
\label{subsec:swarm}

Swarm-based optimization methods model the collective behavior of decentralized agents:
\begin{itemize}
    \item \textbf{Particle Swarm Optimization (\PSO):} Agents (particles) move through the search space influenced by their own best position and the swarm's best position \cite{wang2018particle}.
    \item \textbf{Gray Wolf Optimization (\GWO):} Simulates the hunting behavior of gray wolves with hierarchy-based leadership \cite{mirjalili2014grey}.
\end{itemize}

Both methods can be applied to topology optimization and layout problems but lack explicit diversity mechanisms.

\subsection{Evolutionary Algorithms and Quality-Diversity}
\label{subsec:evolution}

Classical evolutionary algorithms (EAs) search for optimal solutions but suffer from genetic drift in large search spaces. \MapElites \cite{mouret2015illuminating} addresses this by:
\begin{enumerate}
    \item Defining \textit{behavior descriptors} that characterize solutions beyond fitness.
    \item Partitioning the behavior space into niches.
    \item Maintaining an archive of best solutions in each niche.
    \item Enabling simultaneous discovery of diverse, high-quality solutions.
\end{enumerate}

The authors applied both methods and compare the results in this article, showing \MapElites algorithm outperforms classical EAs.

\section{Methodology}
\label{sec:methodology}

\subsection{Wave Function Collapse for Road Networks}
\label{subsec:wfc_method}

Algorithm \ref{alg:wfc} presents the \WFC approach adapted for road networks.

\begin{algorithm}
\caption{Wave Function Collapse (\WFC) for Road Networks}
\label{alg:wfc}
\begin{algorithmic}[1]
\State Initialize grid: all cells can be any valid tile
\While{not all cells collapsed}
    \State Find cell with minimum entropy (fewest possibilities)
    \If{no such cell exists}
        \State \textbf{break}
    \EndIf
    \State Randomly collapse cell to one of its possibilities
    \State Propagate constraints to neighboring cells
    \If{contradiction detected}
        \State Return failure
    \EndIf
\EndWhile
\State \textbf{return} grid
\end{algorithmic}
\end{algorithm}

While \WFC is deterministic given the entropy function, it produces stochastic results through random collapse choices. Therefore, it may fail to find valid solutions, especially on larger grids or with strict constraints.

\subsection{Swarm Algorithms: PSO and GWO}
\label{subsec:swarm_methods}

\subsubsection{Particle Swarm Optimization}
\label{subsubsec:pso}

In \PSO, a particle's velocity is updated as in \cref{eq:part-vel}.

\begin{equation}
v_i^{t+1} = w \cdot v_i^t + c_1 \cdot \text{rand}() \cdot (p_{best}^i - x_i^t) + c_2 \cdot \text{rand}() \cdot (g_{best} - x_i^t)
\label{eq:part-vel}
\end{equation}

where $w$ is the inertia weight, $c_1, c_2$ are cognitive and social coefficients, $p_{best}^i$ is the particle's personal best, and $g_{best}$ is the global best. The position is updated as in \cref{eq:part-pos}.
\begin{equation}
x_i^{t+1} = x_i^t + v_i^{t+1}
\label{eq:part-pos}
\end{equation}

As tile grids are a discrete domain, the authors map continuous velocities to tile placement decisions. This is achieved by converting them to probabilities via the softmax function, as in \cref{eq:pso-softmax}, and finally sampling the tile type $T_{i,y,x}$ from the discrete distribution, as in \cref{eq:pso-sampling}. The fitness function has been applied consistently from \cref{eq:fitness}.

\begin{equation}
    p_{i,y,x,k} = \frac{\exp(v_{i,y,x,k})}{\sum_{j=1}^{K} \exp(v_{i,y,x,j})}
    \label{eq:pso-softmax}
\end{equation}

\begin{equation}
    T_{i,y,x} \sim \text{Categorical}(p_{i,y,x,1}, ..., p_{i,y,x,K})
    \label{eq:pso-sampling}
\end{equation}

\subsubsection{Gray Wolf Optimization}
\label{subsubsec:gwo}

\GWO models predator-prey dynamics. Three tiers of wolves exist: alpha (best), beta (second-best), and omega (worst). Each wolf's position is updated based on the position of alpha, beta and their own:

\begin{equation}
x_i^{t+1} = \frac{1}{3}(x_{\alpha}^t + x_{\beta}^t + x_{\omega}^t) + \epsilon
\end{equation}

where $\epsilon$ is a random perturbation of small amplitude. The \GWO algorithm balances exploration and exploitation through adaptive parameter updates. The fitness function has been applied consistently from \cref{eq:fitness}.

\subsection{Evolutionary Algorithm with MAP-Elites}
\label{subsec:ea_method}

Our primary contribution is an implementation of an \EA with elitism through \MapElites for quality-diversity optimization. Unlike traditional EAs that seek a single optimal solution, \MapElites maintains an archive of solutions distributed across a behavior descriptor space. This allows to explore the search space more extensively than with classical \EA algorithms.

\subsubsection{Algorithm Overview}
\label{subsubsec:ea_overview}

Algorithm \ref{alg:mapelites} outlines our approach.

\begin{algorithm}
\caption{Evolutionary Algorithm with MAP-Elites}
\label{alg:mapelites}
\begin{algorithmic}[1]
\State Initialize empty individual-niche archive: $A = \{\}$
\State Randomize the initial population: $P = \{x_1, \ldots, x_\mu\}$
\For{generation $g = 1, \ldots, G_{\max}$}
    \State Create offspring: $Q = \{\}$
    \For{$l = 1, \ldots, \lambda$}
        \State Select parent $x_p$ from $P$ via tournament selection
        \State Mutate $x_p$ to create offspring $x_c$
        \State Calculate fitness $f(x_c)$ and behavior descriptor $b(x_c)$
        \State Compute niche index: $n = \text{quantize}(b(x_c))$
        \If{$A[n]$ is empty OR $f(x_c) > f(A[n])$}
            \State $A[n] \gets x_c$
        \EndIf
        \State $Q \gets Q \cup \{x_c\}$
    \EndFor
    \State Generate new population with $(\mu, \lambda)$ selection from $Q \cup P$
\EndFor
\State \textbf{return} $A$
\end{algorithmic}
\end{algorithm}

The key difference from standard evolution strategies is the addition of the archive $A$. The authors organized the niche quantization such that the first niche always includes elites carrying the behaviour descriptors of $0$. This is crucial, as some of the behaviour descriptors are unfavorable and thus the authors enforce only $2$ bins to discern behaviors into two niches: favorable (first niche) and unfavorable (second niche).

\subsubsection{Behavior Descriptors}
\label{subsubsec:behavior_descriptors}

We define behavior descriptors to capture diverse structural characteristics of networks:

\begin{enumerate}
    \item \textbf{Connected components}: count of connected components.
    \item \textbf{Cyclomatic complexity}: cyclomatic complexity of the network.
    \item \textbf{Dangling ends}: count of tiles with unconnected edges.
    \item \textbf{Adjacent crossings}: count of adjacent 3+ connectivity tiles.
    \item \textbf{Adjacent turns}: count of adjacent turn tiles.
\end{enumerate}

The behavior space is divided into a fixed number of niches, with each niche storing the best solution discovered within its defined characteristic range. A key contribution of this approach is the quantization of cyclomatic complexity into 25 distinct bins (niches), while all other descriptors are divided into just two niches: one for the value 0, and another for the best individual with a value greater than 0. This design encourages the promotion of individuals that meet the ideal requirement in the first niche, while simultaneously preserving the best individuals in the second niche. The dual-niche structure serves two purposes: firstly it allows for a suboptimal set of individuals to exist temporarily, and secondly it supports both exploration and the eventual generation of individuals that fit the first niche.

\subsubsection{Fitness Function}
\label{subsubsec:fitness}

The fitness function presented in \cref{eq:fitness} balances multiple objectives and is minimized. This fitness function is consistent for all methods assessed in this work.

\begin{equation}
\begin{split}
f(x) = & \ 480 \cdot d(x) + 300 \cdot (c(x) - 1) + 150 \cdot bv(x) + \\ & + 100 \cdot b(x) + 100 \cdot a(x) + 80 \cdot t(x) - 2 \cdot y(x) - 2 \cdot s(x)
\label{eq:fitness}
\end{split}
\end{equation}

where: $c(x)$ is the count of connected components, $d(x)$ is the count of dead-ends (dangling connection edges), $bv(x)$ is the count of tiles being in boundary violation, $b(x)$ is the count of edges, $a(x)$ is the count of adjacent turn tiles, $t(x)$ is the count of adjacent turns, $y(x)$ is the cyclomatic complexity value and $s(x)$ is the quadratic count of consecutive straight roads.

\subsubsection{Mutation Operators}
\label{subsubsec:mutation}

We employ multiple mutation strategies to generate offspring:

\begin{enumerate}
    \item \textbf{Tile change}: Randomly select a tile and change its connectivity to a valid alternative.
    \item \textbf{Crossing insertion}: Randomly select a tile and replace it with any crossing ($\geqslant 3$-connectivity) tile.
\end{enumerate}

The repair mechanism (Section \ref{subsubsec:repair}) ensures that the mutations do not cause invalid offspring.

\subsubsection{Constraint Repair}
\label{subsubsec:repair}

A critical component is the repair algorithm that fixes connectivity mismatches and boundary violations:

\begin{algorithm}
\caption{Connectivity Repair}
\label{alg:repair}
\begin{algorithmic}[1]
\State \textbf{Input:} grid $G$, dirty mask $M$, max iterations $I_{max}$
\State $it$ $\gets 0$
\State changes $\gets$ true
\While{changes AND $it < I_{max}$}
    \State iterations $\gets$ iterations $+ 1$
    \State changes $\gets$ false
    \For{cell $c = (i, j)$ in $M$}
        \For{direction $d$ in $\{N, E, S, W\}$}
            \State Lookup neighbor $t = (n_i, n_j)$ in direction $d$
            \If{$t$ exists}
                \If{cell $c$ and neighbor $t$ have incompatible connectivity at edge $d$}
                    \State Insert compatible tile minimizing changes
                    \State Mark $M[t]$ as dirty
                    \State changes $\gets$ true
                \EndIf
            \Else
                \State Remove connection in direction $d$
            \EndIf
        \EndFor
    \EndFor
\EndWhile
\end{algorithmic}
\end{algorithm}

This repair process keeps the solution valid by enforcing connectivity rules through iterative local adjustments.

\section{Implementation Details}
\label{sec:implementation}

\subsection{Grid Generation Framework}
\label{subsec:framework}

We implemented our methods in Python using the numpy, scipy and OpenCV libraries, with the following core components:

\begin{itemize}
    \item \textbf{Tile management}: Classes encapsulating the entities building up the dataset.
    \item \textbf{Grid evaluation and helpers}: Functions computing all metrics and fitness values, as well as helper functions for performing quantization and other calculations.
    \item \textbf{Constraint checking}: Validation of connectivity, boundary, and crossing constraints.
    \item \textbf{Rendering}: Generation of visual road maps with matching binary masks from tile grids.
\end{itemize}

\subsection{Real-World Data Integration}
\label{subsec:real_world}

The authors collected a low-data dataset of elementary tile elements, photographed in the laboratory setup of the \Duckietown miniature city environment, consisting of dark-colored road tiles with overlaid markings made of adhesive tape striped: yellow for dashed lane separator markings, and red for stop lines. The side lane boundaries are marked with wider white adhesive stripes.

Authors accounted for some redundancy and diversity in the dataset, adding $4$ turn tiles, $4$ straight tiles, $3$ crossings with $3$-way connectivity, and $1$ instance of a crossing with $4$-way connectivity. 
Each tile consists of the following:
\begin{itemize}
    \item RGB image
    \item Binary road mask
    \item Red line mask (stop line markings)
    \item Yellow line mask (lane separator line markings)
\end{itemize}

\begin{figure}
    \centering
    \includegraphics[width=0.65\linewidth]{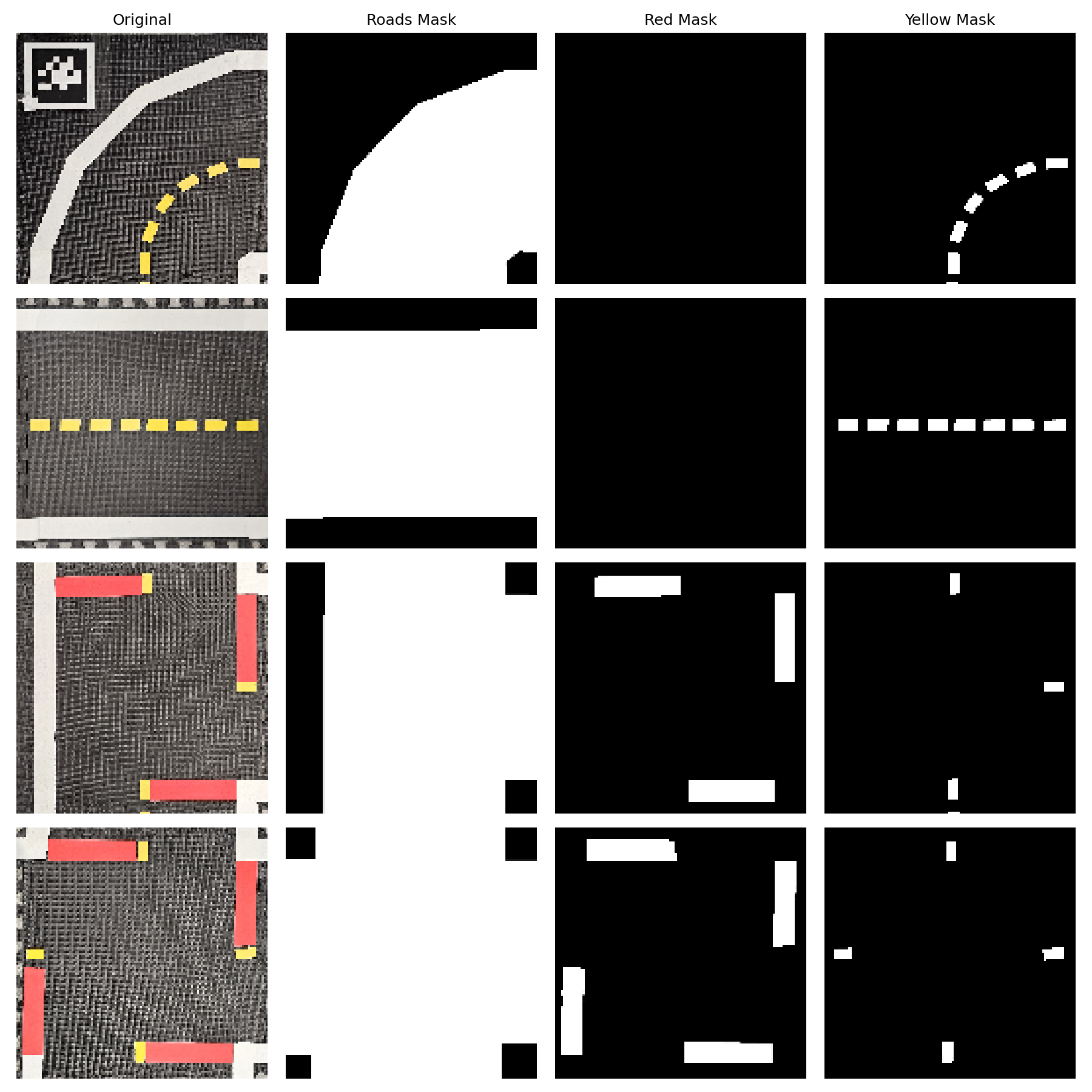}
    \caption{Elementary tile elements with associated semantic segmentation binary masks}
    \label{fig:elementary-tiles}
\end{figure}

All the tiles and masks thereof were augmented by producing $4$ rotated tiles from each. Examples of elementary elements are presented in \cref{fig:elementary-tiles}.

\section{Experimental Results}
\label{sec:results}

\subsection{Experimental Setup}
\label{subsec:exp_setup}

We conducted experiments with the following protocol:

Base configuration for all algorithms is presented in \cref{tab:exp-base-config}.
Algorithm-specific parameters are presented in \cref{tab:exp-alg-config}.

\begin{table}[h]
\centering
\begin{tabular}{lrr}
\toprule
Parameter & Value & Notes \\
\midrule
Grid size & $12 \times 12$ & Tested across scales \\
Mutation rate & 0.3 \\
\bottomrule
\end{tabular}
\label{tab:exp-base-config}
\end{table}

\begin{table}[h]
\centering
\begin{tabular}{lrr}
\toprule
Algorithm & Parameter & Value \\
\midrule
\PSO & Inertia weight & 0.7 \\
 & $c_1$ (cognitive) & 1.5 \\
 & $c_2$ (social) & 1.5 \\
 & Generations & 200 \\
 & Offspring & 40 \\
\midrule
\GWO & Generations & 200 \\
 & Offspring & 40 \\
\midrule
\WFC & Max attempts & 10 \\
 & Entropy threshold & 0 \\
\midrule
\EA & Mutation prob. of tiles & 70\% \\
 & Insertion of crossing prob. & 50\% \\
\bottomrule
\end{tabular}
\label{tab:exp-alg-config}
\end{table}

\begin{enumerate}
    \item \textbf{Baselines}: \WFC, \PSO, \GWO with standard configurations.
    \item \textbf{Proposed method}: \EA with \MapElites, compared with standard \EA.
    \item \textbf{Grid sizes}: $12 \times 12$.
    \item \textbf{Repetitions}: 4 independent runs per configuration.
    \item \textbf{Metrics}: as described in \ref{subsec:metrics}.
\end{enumerate}

\subsection{Quantitative Comparison}
\label{subsec:quant_results}

\begin{figure}
    \centering
    \includegraphics[width=1\linewidth]{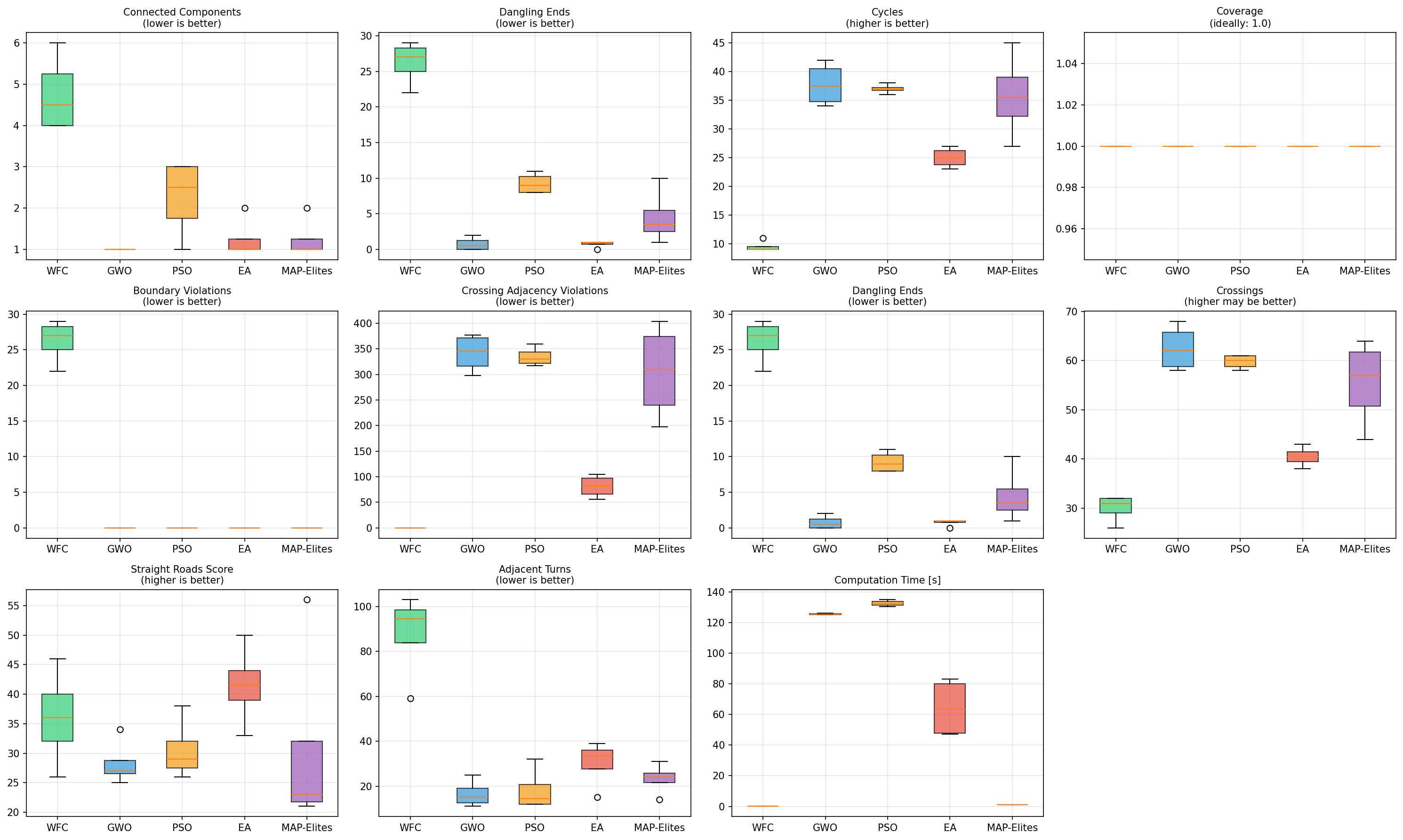}
    \caption{Quantitative comparison of all models, according to metric definitions from \cref{subsec:constraints}}
    \label{fig:quantitative-comparison}
\end{figure}

The quantitative comparison of all models is presented both visually in \cref{fig:quantitative-comparison} and in tabelaric form in: \cref{tab:quantitative-comparison-1}, \cref{tab:quantitative-comparison-2} and \cref{tab:quantitative-comparison-3}.

The results show that:

\begin{enumerate}
    \item \WFC is inferior in all metrics except for crossing adjacency violations (thanks to hard constraints) and computation time (which is $\approx 20-100$ faster than other methods).
    \item The \GWO and \EA algorithms lead in terms of minimizing dangling ends, followed by \MapElites.
    \item While cycles are not definitely better when higher, a reasonably large amount of them may be valuable and such is provided by \EA being in the center, and by \MapElites which is in the higher range, yet provides the widest inter-quartile range (IQR) from all the methods, meaning it provides the largest diversity in this feature.
    \item All methods achieved a $100\%$ coverage.
    \item Boundary violations are only present in \WFC.
    \item Crossing adjacency violations are smallest for \WFC, which has a hard constraint on this metric, and then for the classical \EA; they are comparable for other types of models, but \MapElites has the lower box boundary and lower IQR boundary outperforming \GWO and \PSO.
    \item In terms of crossings, which are neutral or slightly positive, \EA and \MapElites are closest to the average value for all models, with \EA approaching from the low and \MapElites from the high; it must be noted that again, \MapElites has the widest IQR range, meaning it is the most diverse in this matter.
    \item Straight roads score is highest for \EA and \WFC; on the other hand, \MapElites has its upper IQR boundary matching \GWO and \PSO, but also has the widest IQR range.
    \item In terms of adjacent turns, the ones that tend to generate solutions with the lowest amount of them are \GWO and \PSO; yet, both \EA and \MapElites are relatively close and have outlier values in ranges matching \GWO and \PSO means, stating they are able to generate solutions matching \GWO and \PSO in terms of this metric.
    \item In terms of computation time, \WFC is the fastest algorithm followed by \MapElites, then \EA being within the center, finally with \PSO and \GWO being the slowest.
\end{enumerate}

Finally, it can be seen that \WFC is the most 'primitive' and fastest of the solutions, yet it lacks in metrics. \EA is the moderate solution in most metrics and its average scores are often comparable to \MapElites, yet \MapElites usually provides a much wider IQR, meaning it has explored the domain ranges that \EA has not and therefore provides a larger diversity of solutions. \GWO and \PSO might seem comparable to \EA and \MapElites in some cases, yet they firstly lose in terms of straight roads score or crossing adjacency violations and secondly, they offer the smallest diversity of solutions.
         
\begin{table}
    \centering
    \begin{tabular}{lrrrrrr}
    \toprule
     & \multicolumn{2}{c}{CO} & \multicolumn{2}{c}{DE} & \multicolumn{2}{c}{CY} \\
     & $\mu$ & $\sigma$ & $\mu$ & $\sigma$ & $\mu$ & $\sigma$ \\
    Method &  &  &  &  &  &  \\
    \midrule
    EA & 1.25 & 0.5 & 0.75 & 0.5 & 25 & 1.826 \\
    GWO & 1 & 0 & 0.75 & 0.957 & 37.75 & 3.862 \\
    MAP-E & 1.25 & 0.5 & 4.5 & 3.873 & 35.75 & 7.455 \\
    PSO & 2.25 & 0.957 & 9.25 & 1.5 & 37 & 0.816 \\
    WFC & 4.75 & 0.957 & 26.25 & 3.096 & 9.5 & 1 \\
    \bottomrule
    \end{tabular}
    \caption{Quantitative measures table 1 of 3, where: CO - connected components; DE - dangling (dead) ends; CY - cyclomatic complexity}
    \label{tab:quantitative-comparison-1}
\end{table}

\begin{table}
    \centering
    \begin{tabular}{lrrrrrr}
    \toprule
     & \multicolumn{2}{c}{BV} & \multicolumn{2}{c}{ACV} & \multicolumn{2}{c}{CV} \\
     & $\mu$ & $\sigma$ & $\mu$ & $\sigma$ & $\mu$ & $\sigma$ \\
    Method &  &  &  &  \\
    \midrule
    EA & 0 & 0 & 81.25 & 22.66 & 1 & 0 \\
    GWO & 0 & 0 & 341.8 & 38.06 & 1 & 0 \\
    MAP-E & 0 & 0 & 305.2 & 95.66 & 1 & 0 \\
    PSO & 0 & 0 & 334.5 & 19.16 & 1 & 0 \\
    WFC & 26.25 & 3.096 & 0 & 0 & 1 & 0 \\
    \bottomrule
    \end{tabular}
    \caption{Quantitative measures table 2 of 3, where: BV - boundary violations; ACV - adjacent crossing violations scaled by connectivity; CV - coverage}
    \label{tab:quantitative-comparison-2}
\end{table}

\begin{table}
    \begin{tabular}{lrrrrrr}
    \toprule
     & \multicolumn{2}{c}{CR} & \multicolumn{2}{c}{SR} & \multicolumn{2}{c}{AT} \\
     & $\mu$ & $\sigma$ & $\mu$ & $\sigma$ & $\mu$ & $\sigma$ \\
    Method &  &  &  &  &  &  \\
    \midrule
    EA & 40.5 & 2.082 & 41.5 & 6.952 & 30.25 & 10.56 \\
    GWO & 62.5 & 4.796 & 28.25 & 3.948 & 16.5 & 6.191 \\
    MAP-E & 55.5 & 8.963 & 30.75 & 16.88 & 23.25 & 6.994 \\
    PSO & 59.75 & 1.5 & 30.5 & 5.26 & 18.25 & 9.465 \\
    WFC & 30 & 2.828 & 36 & 8.327 & 87.75 & 19.69 \\
    \bottomrule
    \end{tabular}
    \caption{Quantitative measures table 3 of 3, where: CR - crossings count; SR - quadratic consecutive straight roads length; AT - adjacent turns count}
    \label{tab:quantitative-comparison-3}
\end{table}

\subsection{Visual Results}
\label{subsec:visual_results}

\begin{figure}
    \centering
    \includegraphics[width=\linewidth]{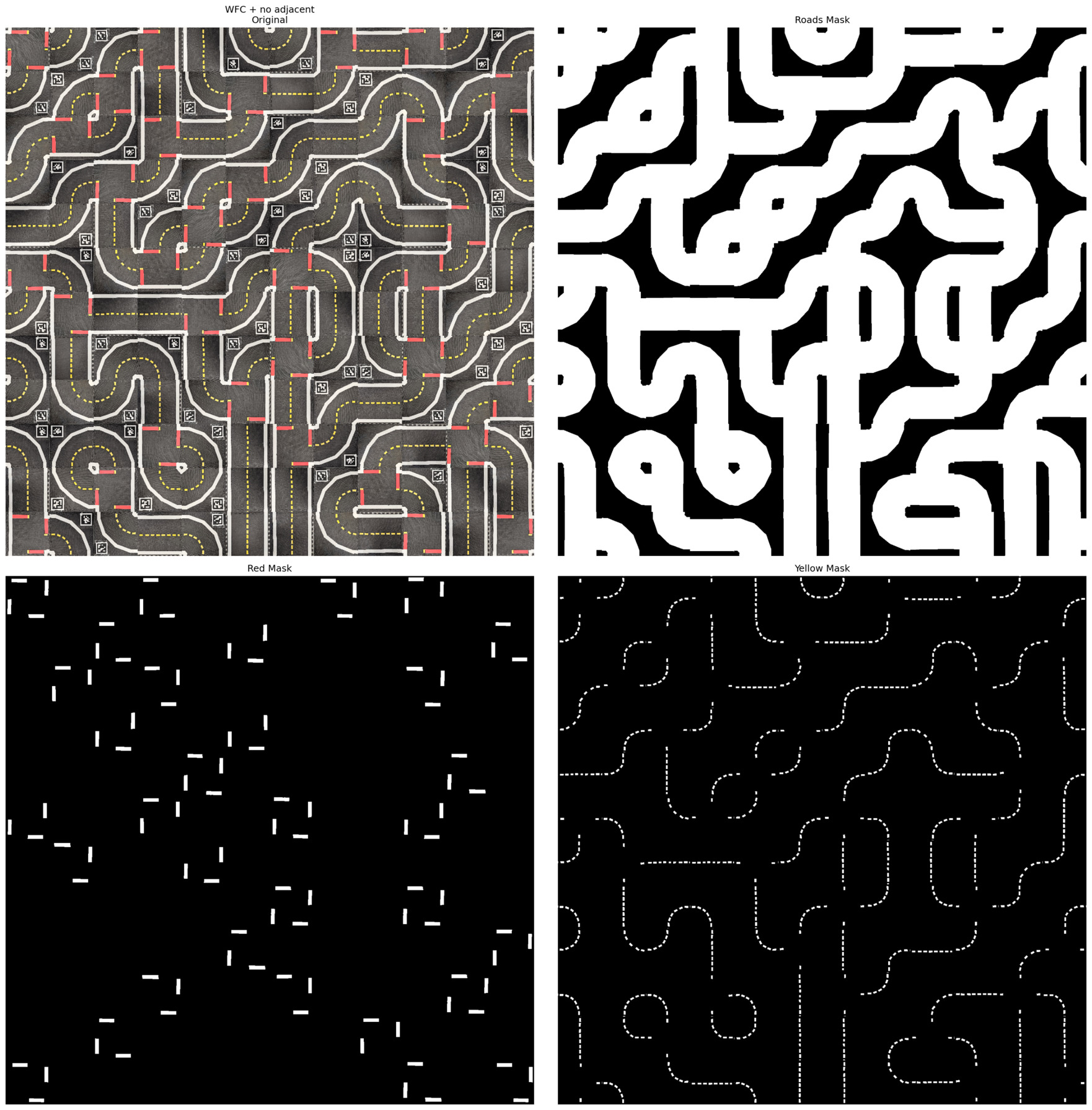}
    \caption{Example $14 \times 14$ output generated by the \WFC algorithm}
    \label{fig:wfc-14x14-out}
\end{figure}

Additionally, for empirical presentation, example results are presented as follows for a single run for a grid of size $14 \times 14$ and parameters as stated in \cref{tab:exp-base-config} and \cref{tab:exp-alg-config}.
The \WFC algorithm provides results as in \cref{fig:wfc-14x14-out}, which exhibit a pretty chaotic way of routing the roads, as well as the network is split into $3$ components. It is also visible that the network has little redundancy, having $85$ cut-edges. The solution has $32$ dead ends, resulting in a low quality. On the other hand, the quadratic running length score of straight roads is $46$.

\begin{figure}
    \centering
    \includegraphics[width=\linewidth]{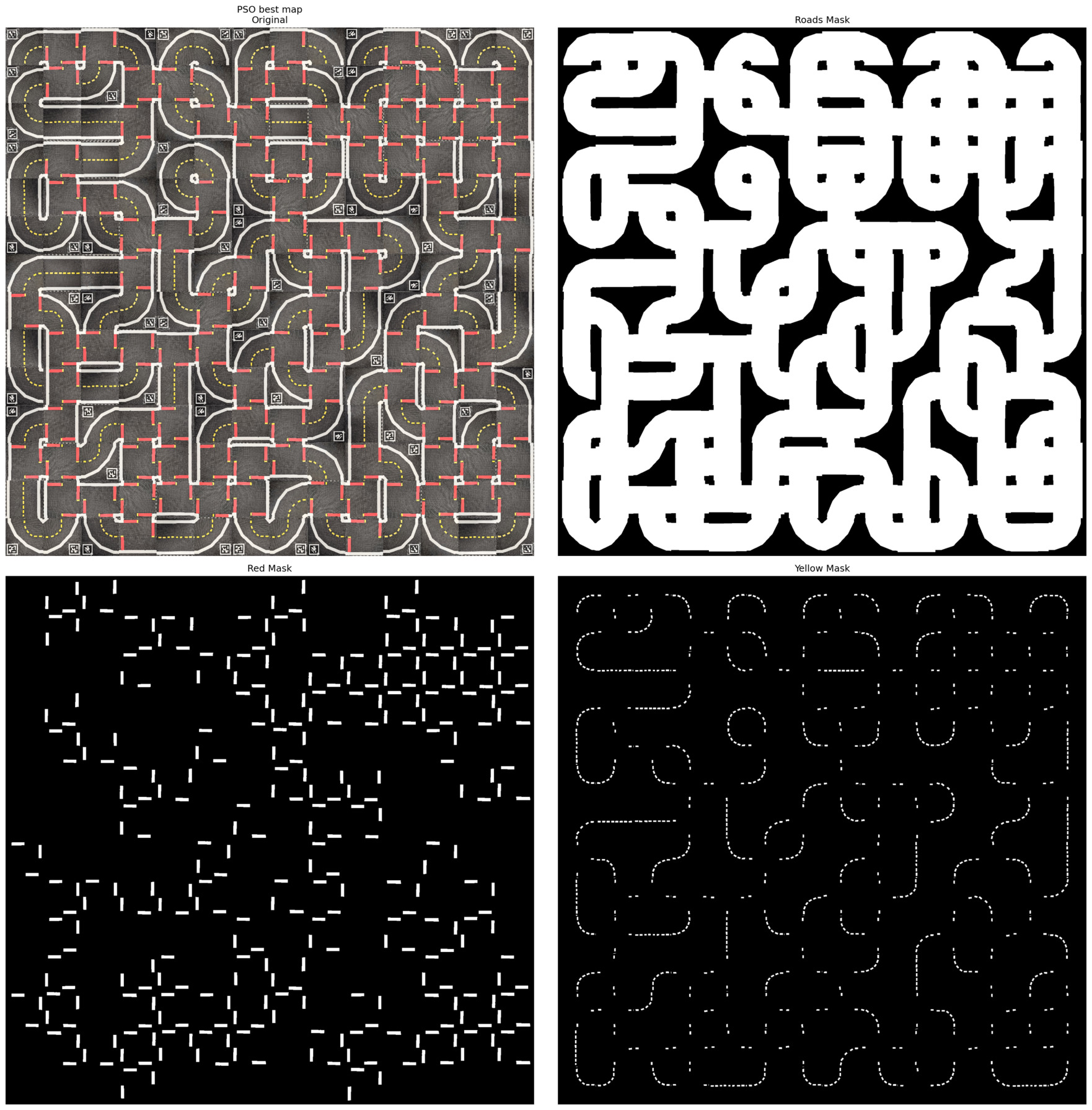}
    \caption{Example $14 \times 14$ output generated by the \PSO algorithm}
    \label{fig:pso-14x14-out}
\end{figure}

The \PSO algorithm provides results as in \cref{fig:pso-14x14-out}, which exhibit $6$ dangling ends, $0$ boundary violations, $11$ bridges, a $25$ quadratic running length score of straight roads, but on the other hand it is clearly visible that a significant part of the network is crossings, which are adjacent to each other, producing an improbable layout and a connectivity-scaled adjacent crossing violation score of $607$.

\begin{figure}
    \centering
    \includegraphics[width=\linewidth]{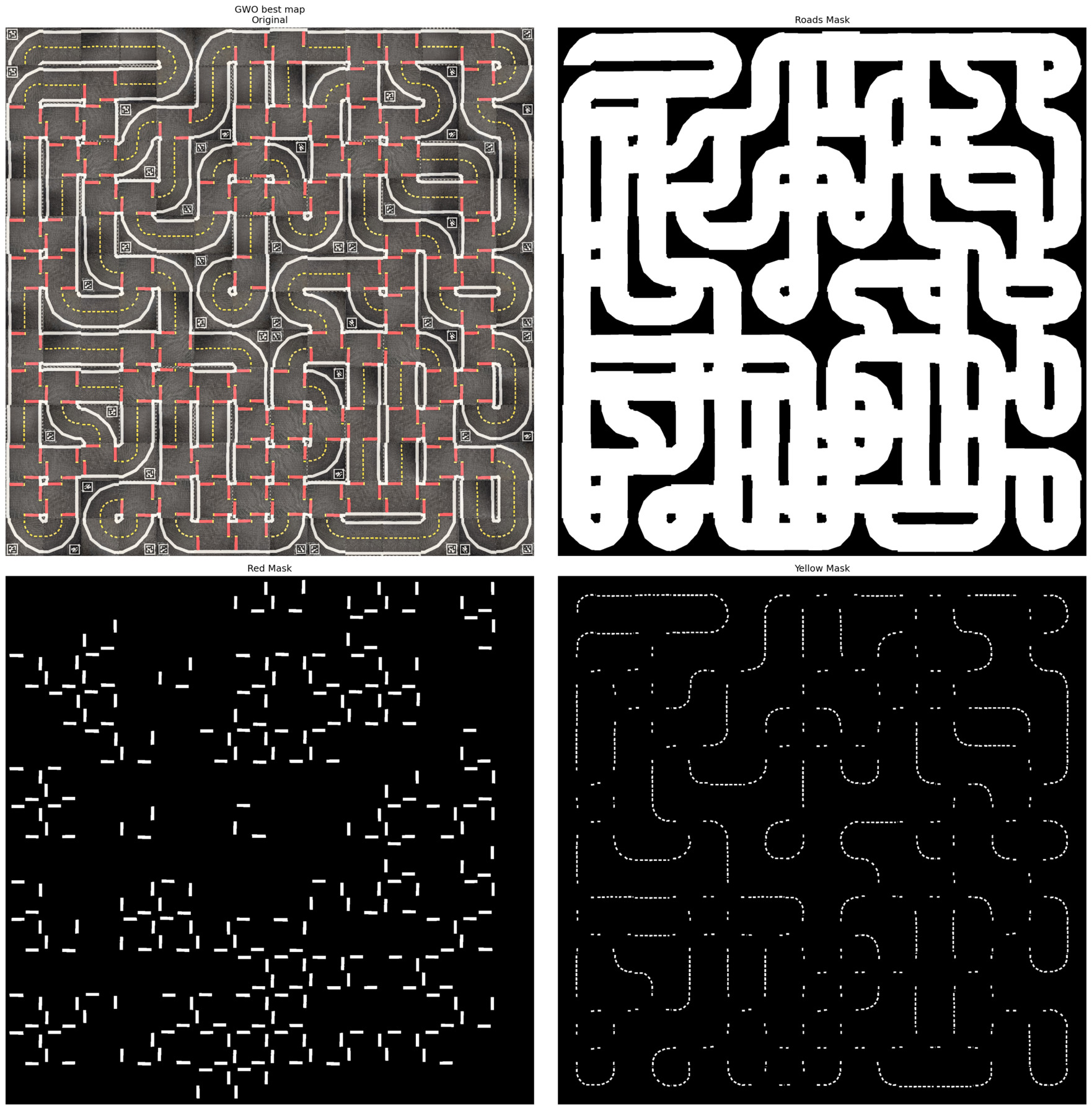}
    \caption{Example $14 \times 14$ output generated by the \GWO algorithm}
    \label{fig:gwo-14x14-out}
\end{figure}

The \GWO algorithm provides results as in \cref{fig:gwo-14x14-out}, which exhibit $1$ dangling end, $0$ boundary violations, $11$ bridges, a $57$ quadratic running length score of straight roads, but also a connectivity-scaled adjacent crossing violation score of $537$.

\begin{figure}
    \centering
    \includegraphics[width=\linewidth]{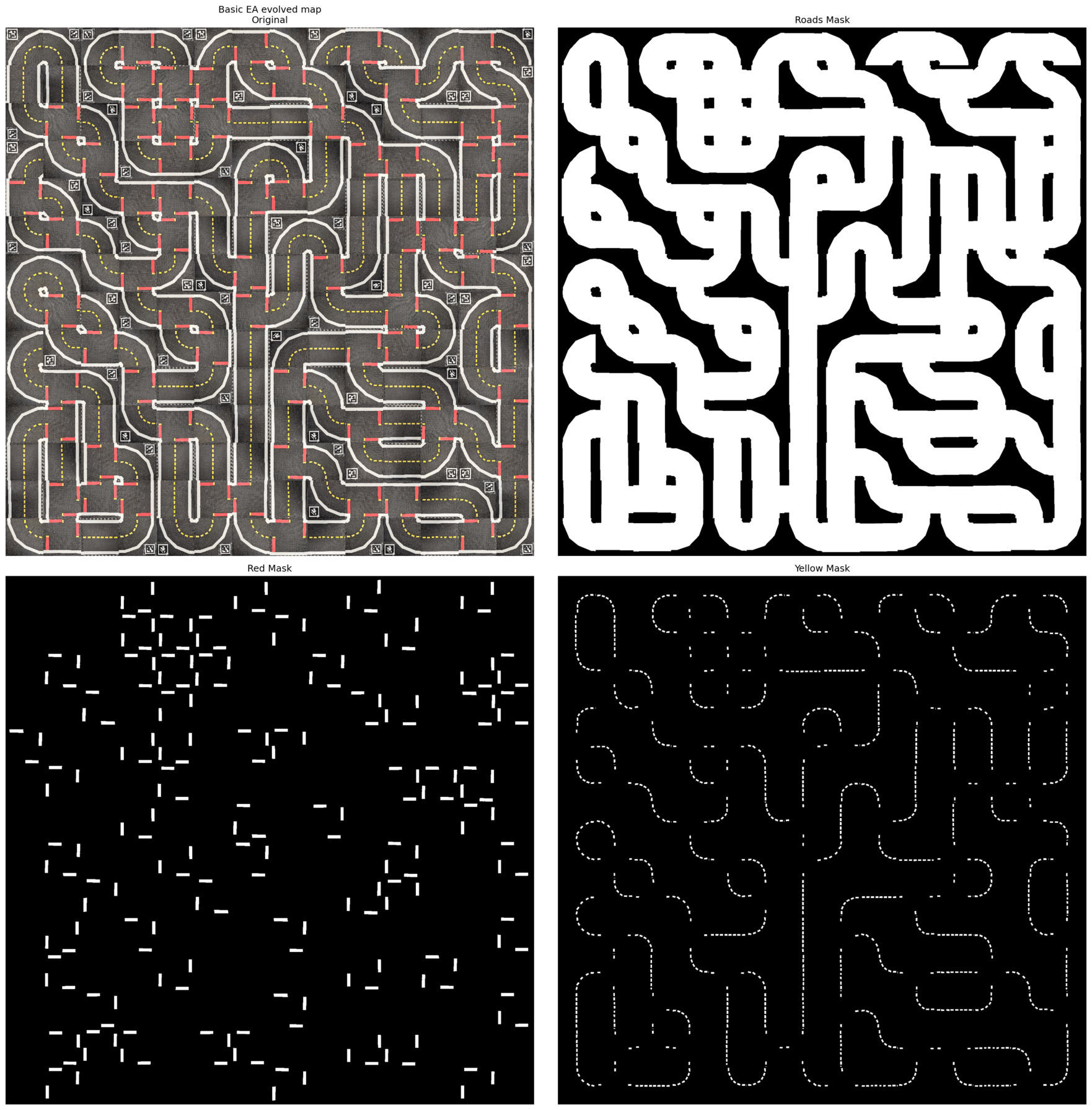}
    \caption{Example $14 \times 14$ output generated by the \EA algorithm}
    \label{fig:ea-14x14-out}
\end{figure}

The proposed \EA algorithm provides results as in \cref{fig:ea-14x14-out}, which exhibit $5$ dangling ends, $15$ cut-edges and visually present consecutive sharp turns in the left half of the image.

\begin{figure}
    \centering
    \includegraphics[width=\linewidth]{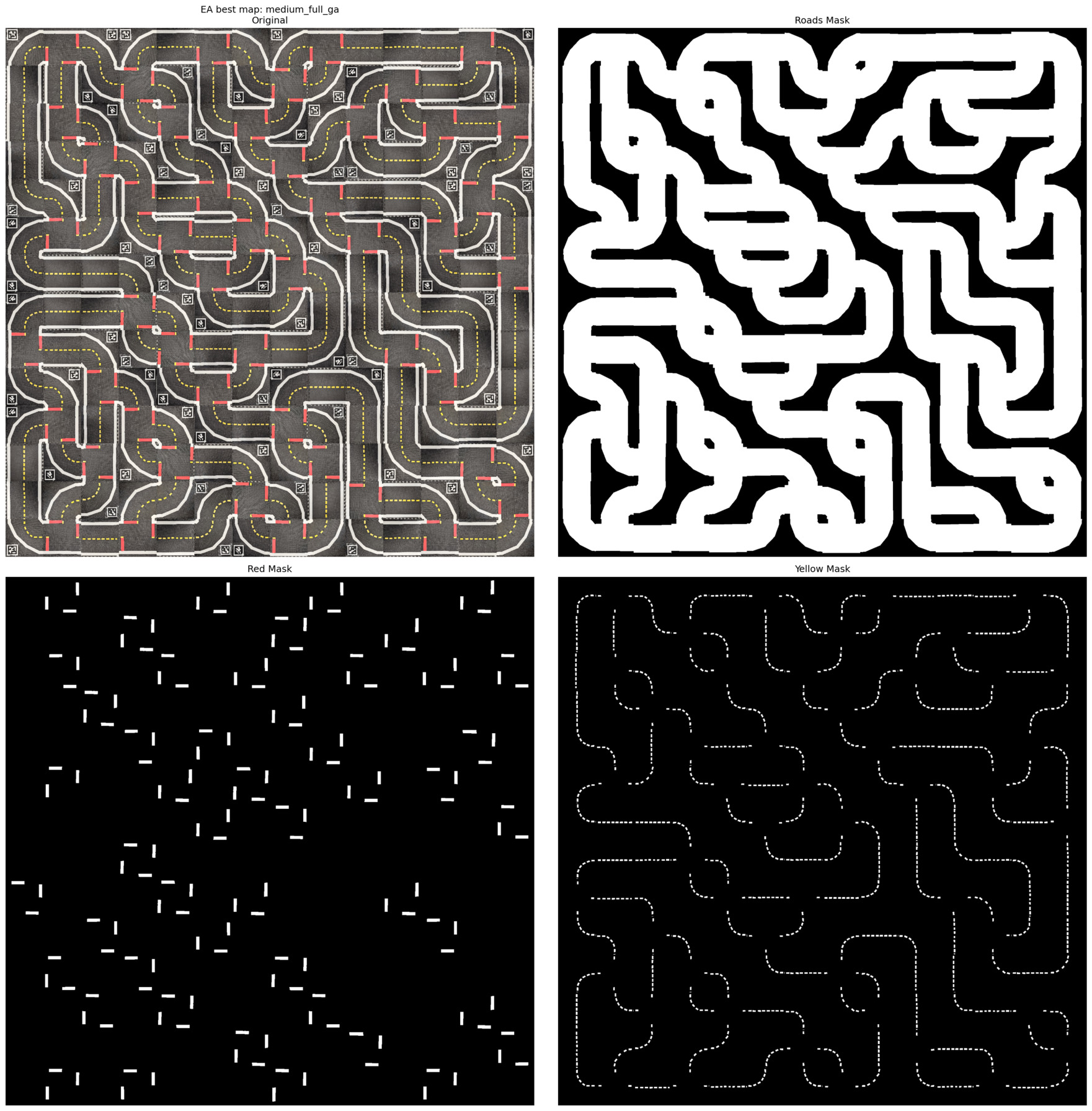}
    \caption{Example $14 \times 14$ output generated by the \MapElites algorithm}
    \label{fig:mapelites-14x14-out}
\end{figure}

The proposed \MapElites algorithm provides results as in \cref{fig:mapelites-14x14-out}, which exhibit $0$ boundary and adjacent crossing violations, $0$ dead ends, $0$ bridges, which is crucial for redundant connectivity, thus outperforming all others.

\section{Conclusions}
\label{sec:conclusion}

We have presented a comprehensive comparative study of road network generation methods, with a focus on realistic, constrained small-scale networks.

The proposed approach enables the creation of diverse, realistic synthetic datasets suitable for training segmentation models or navigation algorithms. The explicit diversity maintenance via \MapElites allows researchers to explore the solution space more extensively, discovering individuals that may not be found by other methods and effectively greatly boosting the metrics of the classical \EA approach.

\section{Limitations and Future Research}

As stated in the introduction, this research concentrated on generating networks that resemble predefined reality rules proposed by the authors and by generating the new dataset using real-world imagery of elementary elements. Future research may definitely assess the impact of the reality gap phenomenon on semantic segmentation models trained on the dataset, which the authors plan on following up on.

\section{Acknowledgements}

According to ACM guidelines, authors note that Large Language Models (LLMs) have been used in linguistic proofreading and corrections, as well as for code assist and autocomplete during the implementation of the described algorithms.

\bibliographystyle{plain}
\bibliography{software}

\end{document}